# ROBUST BUILDING FOOTPRINT EXTRACTION FROM BIG MULTI-SENSOR DATA USING DEEP COMPETITION NETWORK

M. Khoshboresh Masouleh *, M.R. Saradjian

School of Surveying and Geospatial Engineering, College of Engineering, University of Tehran, Tehran, Iran - (m.khoshboresh, sarajian)@ut.ac.ir

**ABSTRACT:**

Building footprint extraction (BFE) from multi-sensor data such as optical images and light detection and ranging (LiDAR) point clouds is widely used in various fields of remote sensing applications. However, it is still challenging research topic due to relatively inefficient building extraction techniques from variety of complex scenes in multi-sensor data. In this study, we develop and evaluate a deep competition network (DCN) that fuses very high spatial resolution optical remote sensing images with LiDAR data for robust BFE. DCN is a deep superpixelwise convolutional encoder-decoder architecture using the encoder vector quantization with classified structure. DCN consists of five encoding-decoding blocks with convolutional weights for robust binary representation (superpixel) learning. DCN is trained and tested in a big multi-sensor dataset obtained from the state of Indiana in the United States with multiple building scenes. Comparison results of the accuracy assessment showed that DCN has competitive BFE performance in comparison with other deep semantic binary segmentation architectures. Therefore, we conclude that the proposed model is a suitable solution to the robust BFE from big multi-sensor data.

## 1. INTRODUCTION

In recent years, robust automated algorithm development for the extraction of building footprints from remotely sensed data is a hot topic for research and commercial projects [1]. In practice, there are two issues that are essential in building footprint extraction (hereafter called BFE for short). First, data source selection that plays an important role in information extraction. Second, the appropriate knowledge such as deep learning (DL) for accurate and efficient data processing

### 1.1 Data Source

The main factors of data source selection for BFE are related to separation between buildings from non-buildings (spatial resolution considerations), the confusing effect of vegetation-cover on building detection (spectral resolution considerations), and the shade of building/non-building objects and lighting conditions (types of sensors considerations).

The majority of related works that uses the multi-sensor data consist of very high spatial resolution multispectral images and light detection and ranging (LiDAR) data [2]–[6]. LiDAR data (also known as point clouds) and digital surface models (DSMs) generated by aerial platform equipped with airborne laser scanning, such as unmanned aerial vehicle or aircraft are applicable for the automatic BFE, because these data provide the geometrical features of buildings shapes [7]–[10]. Moreover, fusion of LiDAR point clouds and very high spatial resolution multispectral images offers an efficient data source for BFE [11]. Hence, on the basis of the fresh information mentioned above, still, the use of these combined data can be a very convenient source of data for BFE and also has many key issues in processing unresolved, particularly suited to big multi-sensor data. Since the big multi-sensor data has a massive volume of geospatial aerial or satellite data, it is extremely difficult or impossible to process using traditional algorithms [12], [13].

### 1.2 Deep Learning

In the past few years, DL approaches play a crucial role in analysing the big image data [14]–[17]. The DL approaches incorporate two influential concepts in an optimal big data analysis workflow for image data [18]. Sparse topological connectivity in convolutional neural networks which is the most common type of DL and weight sharing across deep network can be used to improve the generalization in DL. DL allows computational algorithms of complex structures that are fused of various processing layers to learn features of input image with various levels of abstraction [19].

In remote sensing image processing, DL approaches have been widely used in classification of the land-use and land-cover using low-resolution images [20], segmentation and object detection from high-resolution image [21], and extracting and interpreting ambiguous information from single image, such as DSM [22]. Moreover, due to difficult challenges of extracting building from remotely sensed data, BFE is one of the most important objectives in geomatics science. Table 1 presents the overview of the methods recently published in applied DL for BFE, based on highlighting characteristics with focus on research innovations. Although the related methods are powerful, but it's not still an outstanding performance for BFE, particularly, in robust BFE from big image data.

---

\* Corresponding author

**Table 1.** Overview of recent BFE researches using DL models

| Algorithm | Highlight of characteristics |
|---|---|
| ConvNet+ SignedDist [23] | - Integrating multi-layer information and a unique output representation<br>- Combine signed-distance labels with ConvNet |
| ABF+SegNet [16] | - Fusion of convolutional layer with adaptive bilateral filter<br>- The minimum bounding rectangle were used for outline regularization |
| SegNet-Dist-Fused [6] | - Combine signed-distance labels with SegNet<br>- No requirement of post-processing |
| Res-U-Net [24] | - Feature extraction based on several residual blocks<br>- A concatenation with the corresponding block from the encoding part is designed |
| MC–FCN [18] | - A bottom-up / top-down multi-constraint fully convolutional network<br>- Basic structure based on fusion of U-Net and three extra multi-scale constraints |
| GRRNet [2] | - Encoding stage based on residual learning network<br>- Improving feature learning with a gated feature labelling unit |

In this study, we focus on the key challenges for creating robust BFE model. In this regard, the major contributions of this study to the robust BFE from big multi-sensor data using DL algorithms are as follows:

- For the architectural structure, an efficient deep competition network (DCN) is proposed based on the encoder vector quantization with classified structure and superpixel for BFE from big multi-sensor data using very high spatial resolution multispectral images and LiDAR data.

- In the feature learning step, the very high spatial resolution multispectral image superpixel-based features of big multi-sensor data are combined with LiDAR data (*i.e.* DSM) superpixel-based features in variety of complex roof shapes and textures in large urban areas.

This paper is organized as follows: section (2), provides essential context around the proposed method. The experiment results on big multi-sensor data are demonstrated in section (3). Insight discussion on gains from the study is presented in final section.

## 2. METHODOLOGY

In this study, two semantic segmentation methods based on DL architectures, including Res-U-Net [24], and ABF+SegNet [16], have been used for BFE results comparisons. The models have been selected because of their good performance in BFE from multi-sensor data.

### 2.1 Res-U-Net

Res-U-Net is a new fully convolutional network for semantic binary segmentation. The architecture uses the modified versions of U-Net and ResNet [24]. This model features a robust encoder-decoder structure for BFE, because upsampling features in the decoder block and the corresponding max-pooling features in encoder block are made separately and concatenated for other upsampling layers [2].

### 2.2 ABF+SegNet

Khoshboresh Masouleh and Shah-Hosseini (2018) proposed a fusion-based architecture, called ABF+SegNet for building outline enhancement using remote sensing big image data, where the SegNet model [25] acts as the high-level features generator based on adaptive bilateral filter. ABF+SegNet achieves excellent performance on RGB images for building extraction.

### 2.3 Deep Competition Network (DCN)

In this paper, we proposed an efficient DCN architecture for BFE from big multi-sensor data. DCN is a deep superpixelwise convolutional encoder-decoder architecture using the encoder vector quantization [26] with classified structure for semantic binary segmentation. Our proposed DCN consists of five encoding-decoding blocks with convolutional weights for robust binary representation (feature) learning.

Figure 1 shows the processing chain of the proposed algorithm. In this algorithm, a superpixel segmentation method called Simple Linear Iterative Clustering - SLIC [27], which is one of the mostly used image segmentation algorithms is used to generate basic processing unit.

The competition function to BFE in DCN can be described as follows:

$$n = \arg\min_i \{ \| \frac{1}{2} \sum f(I-O)^2 \| \} \qquad (1)$$

where  $n$ = winner index
  $i$ = binary value
  $f$ = sigmoid function

$I$ = input
$O$ = prediction (output)

In Equation 1, a loss function (*e.g.,* sigmoid) is defined monotonically increasing as follows:

$$f(I) = \frac{1}{1+e^I} \qquad (2)$$

where  $f$ = sigmoid function
$e$ = Napier's constant (= 2.7182)

In each block, batch normalization function [28], and dropout-based regularization technique have been used to improve performance in training stage with focus on reducing overfitting [29]. The activation function on this model is the rectified linear unit (ReLU) function [19] and the proposed competition function appears in the final output layers. Batch normalization, and ReLU functions are computed as follows, respectively:

$$I_{BN} = \frac{I_i - B_m}{B_v} \qquad (3)$$

$$ReLU(I_i) = \max(0, I_i) = \begin{cases} I_i, & \text{if } I_i \geq 0 \\ 0, & \text{if } I_i < 0 \end{cases} \qquad (4)$$

where  $I_i$ = input
$B_m$ = batch mean
$B_v$ = batch variance

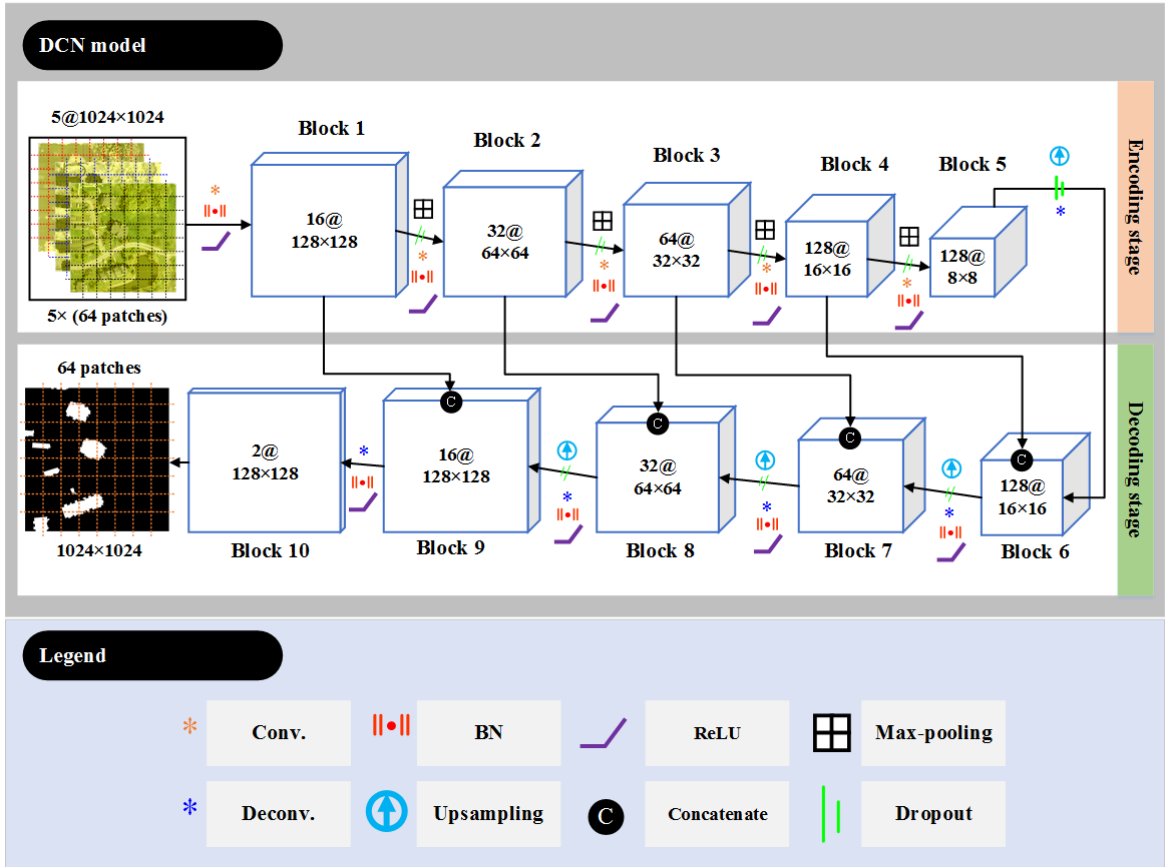

Figure 1. Architecture of the proposed DCN

### 3. EXPERIMENTS AND RESULTS

#### 3.1 Big Multi-Sensor Data

The experiments on big multi-sensor data consisting of:
(1) Very high spatial resolution multispectral images with the four spectral bands (*i.e.* red, green, blue, and NIR) and a ground sampling distance (GSD) of 0.5 foot obtained from the State of Indiana in the US with multiple building scenes,
(2) DSM generated from LiDAR point clouds with a GSD of 0.5 foot by the Indiana Office of Information Technology,
(3) The normalized difference vegetation index (NDVI) generated from red and NIR bands, and
(4) OpenStreetMap shapefiles used as ground truth map to validate results.

Figure 2 displays the research site of the big multi-sensor data. The research site covers about 950 km² from the Indianapolis city in the US. Moreover, big multi-sensor data consist of RGB, DSM, NDVI and building footprint map without major misalignment in the projection of the North American Datum (NAD). Table 2 shows the splitting statistics of the big multi-sensor data.

**Table 2.** Dataset splitting statistics

|            | Tiles |
|------------|-------|
| Training   | 256   |
| Validation | 40    |
| Testing    | 3     |
| Total      | 299   |

For more information about the coordinate system of data, please see (http://gis.iu.edu/datasetInfo/statewide/in_2011.php). In Figure 2, the orange boundary indicates the study area.

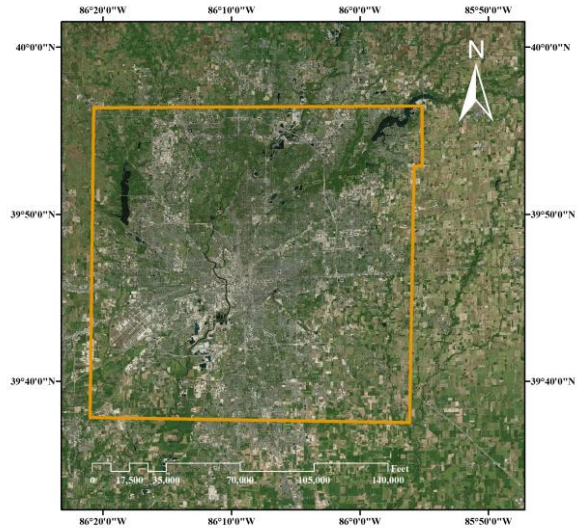

Figure 2. Location of study area

### 3.2 DCN Implementation

DCN was implemented using Keras [30] on the free cloud Tesla K80 GPU and 12GB of RAM in google Colarboratory (Colab), but there is not enough RAM for big data storage. For this reason, we integrated google drive (free cloud storage) with Colab for memory enhancing. Keras is a powerful and open source DL framework written in Python. DCN was trained and validated with adaptive moment estimation (ADAM) optimizer using the default parameters [31] and with a batch size of 64 for 250 epochs for BFE. Moreover, 299 tiles from Indiana each of the size 1024×1024 pixels are processed using a 128×128 pixels sliding window in order to reduce memory consumption.

### 3.3 Building Footprint Extraction

As shown in Figures 3-5, three representative samples are selected from the test area with densely distributed buildings for assessing the performance of the baseline models in comparison with our proposed model. The evaluation metrics of overall accuracy (OA) and intersection over union (IoU) are used to evaluate the performance of the models based on the values of true positive (white), false positive (red), false negative (blue), and true negative (black). The OA and IoU are the most used evaluation measures for BFE in many previous studies [5], [6], [15], [24], [32], [33]. The OA and IoU are defined as:

$$OA = \frac{TP + TN}{TP + FP + FN + TN} \quad (2)$$

$$IoU = \frac{TP}{TP + FN + FP + e} \quad (3)$$

where  *TP* = true positive  
 *FP* = false positive  
 *FN* = false negative  
 *TN* = true negative  
  *e* = 10⁻¹⁵ (to avoid division by zero)

In Table 3 we compared DCN's performance with Res-U-Net and ABF+SegNet models for three test samples. Bold fonts denote the best results and the underlined fonts denote the second best results. The DCN model's OA reaches 94%, 99% and 99% while the IoU reaches the 93%, 99% and 98% for each test samples, respectively. The results demonstrated that DCN model is a suitable solution to robust BFE from big multi-sensor data, especially on the different types of roofs.

**Table 3.** Quantitative evaluation on the test area

|  | Model | Res-U-Net | ABF+SegNet | DCN |
|---|---|---|---|---|
| Case-1 | OA | 91% | 91% | **94**% |
|  | IoU | 89% | 90% | **93**% |
| Case-2 | OA | **99**% | **99**% | **99**% |
|  | IoU | 98% | 98% | **99**% |
| Case-3 | OA | 96% | 97% | **99**% |
|  | IoU | 91% | 96% | **98**% |

### 3.4 Computational Cost Analysis

Computational cost is an important factor in big data processing, particularly for real world applications such as BFE, because hardware limitations (*e.g.* memory consumption, processing system, etc.) in the real world. Therefore, selection of appropriate computational space for big data processing with optimal computational cost is necessity. Cloud computing is an internet-based space for reducing the computational cost in DL experiments. In this paper, we used Colab (cloud computing platform) for training efficiency and computational cost analysis. For this purpose, we trained all models (Res-U-Net, ABF+SegNet, and DCN) based on big multi-sensor data in Colab. The evaluation metric for computational cost analysis is defined as [34]:

$$CC = \frac{NE \times TT}{60} \quad (4)$$

where   $CC$ = computational cost (in min)
$NE$ = number of epochs
$TT$ = training time per epoch (in sec)

The computational cost results are presented in Table 4. Bold font denotes the best result and the underlined font denotes the second best result.

**Table 4.** Computational cost results on the test area

| Model | Res-U-Net | ABF+SegNet | DCN |
|---|---|---|---|
| CC (min) | 535 | 590 | **461** |

### 4. CONCLUSION

In this paper, we focus on tackling the regularization of building outlines problem in very high spatial resolution remote sensing images by proposing a model based on DL and superpixel segmentation called DCN. Most important feature of this model is exploitation of vector quantization theory and convolutional layers in creating a DL network for BFE. In order to train the proposed model, INDIANA dataset was used. Results of applying the proposed method on three test samples indicate improvement in training speed and increase in accuracy and validity of BFE from big multi-sensor data that contains very high spatial resolution multispectral images and LiDAR data. DCN model, automatically extracts the buildings from input data based on the encoder vector quantization framework (supervised learning). In order to evaluate the results, we compared our model with two powerful DL models. Based on the statistical results shown in Table 3, the accuracy is somewhat better, but the IoU (a scale invariant metric) is obviously improved. Future studies can be conducted to increase performance of DCN through optimizing network depth and improving superpixel segmentation methods to reinforce BFE.

### ACKNOWLEDGEMENTS

The INDIANA dataset were obtained freely from the Indiana Office of Information Technology. We thank the Indiana Office of Information Technology. Finally, we would like to thank the anonymous referee for his/her helpful comments.

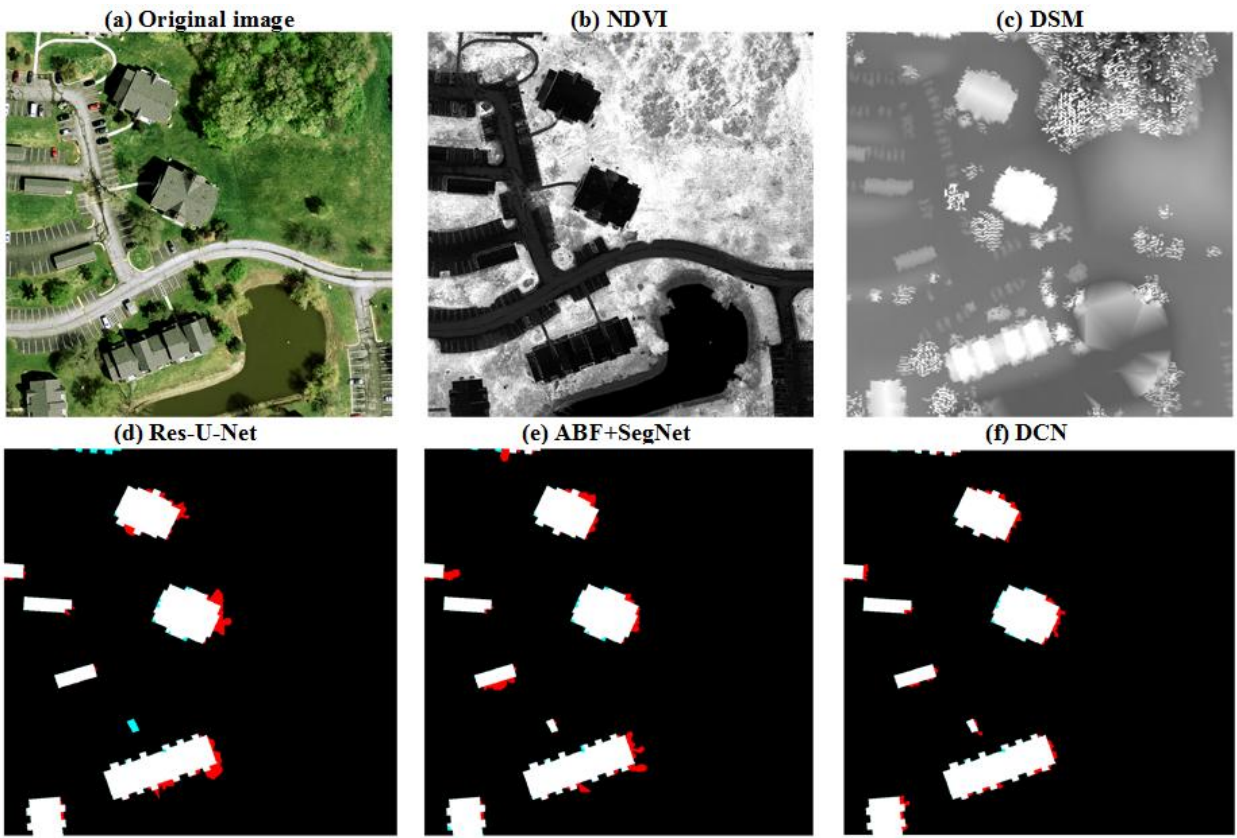

Figure 3. Comparison of BFE results using Res-U-Net, ABF+SegNet and DCN with ground truth in Case-1

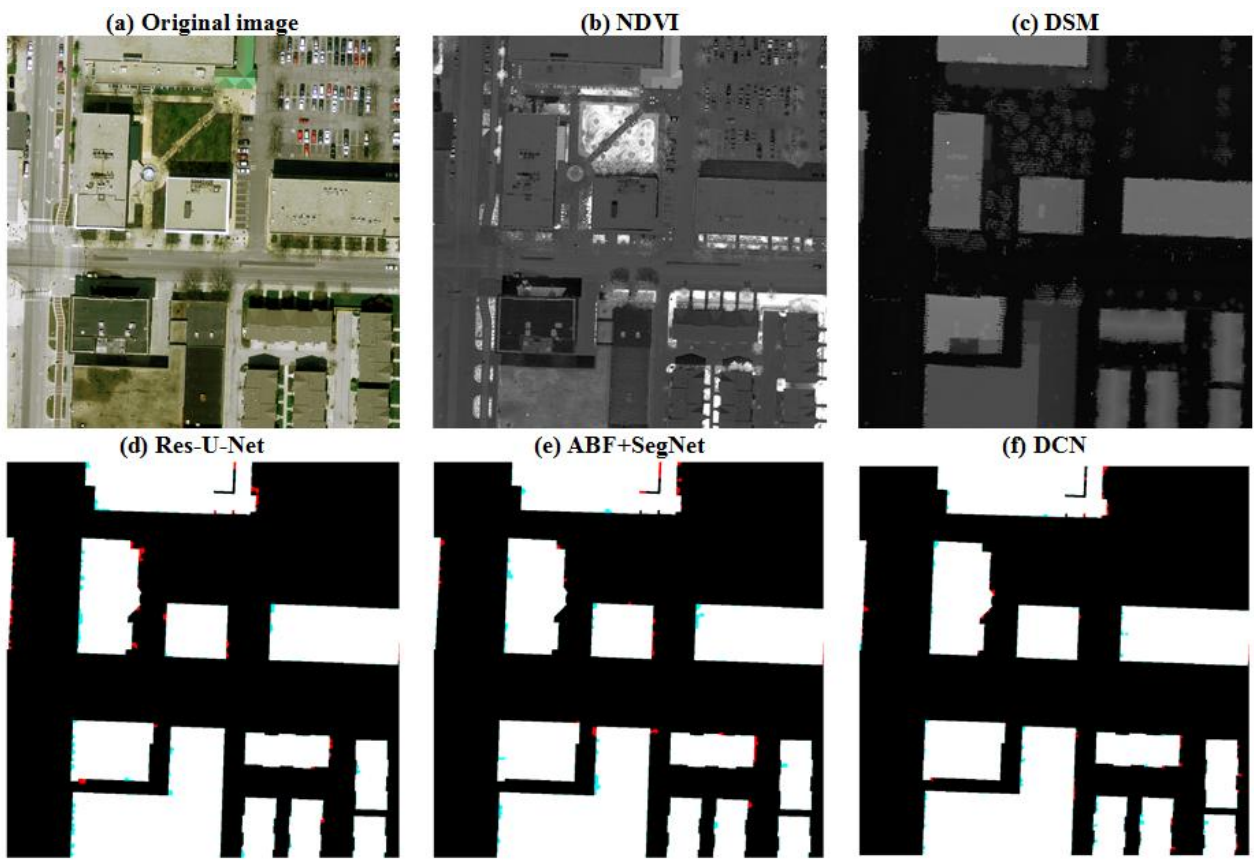

Figure 4. Comparison of BFE results using Res-U-Net, ABF+SegNet and DCN with ground truth in Case-2

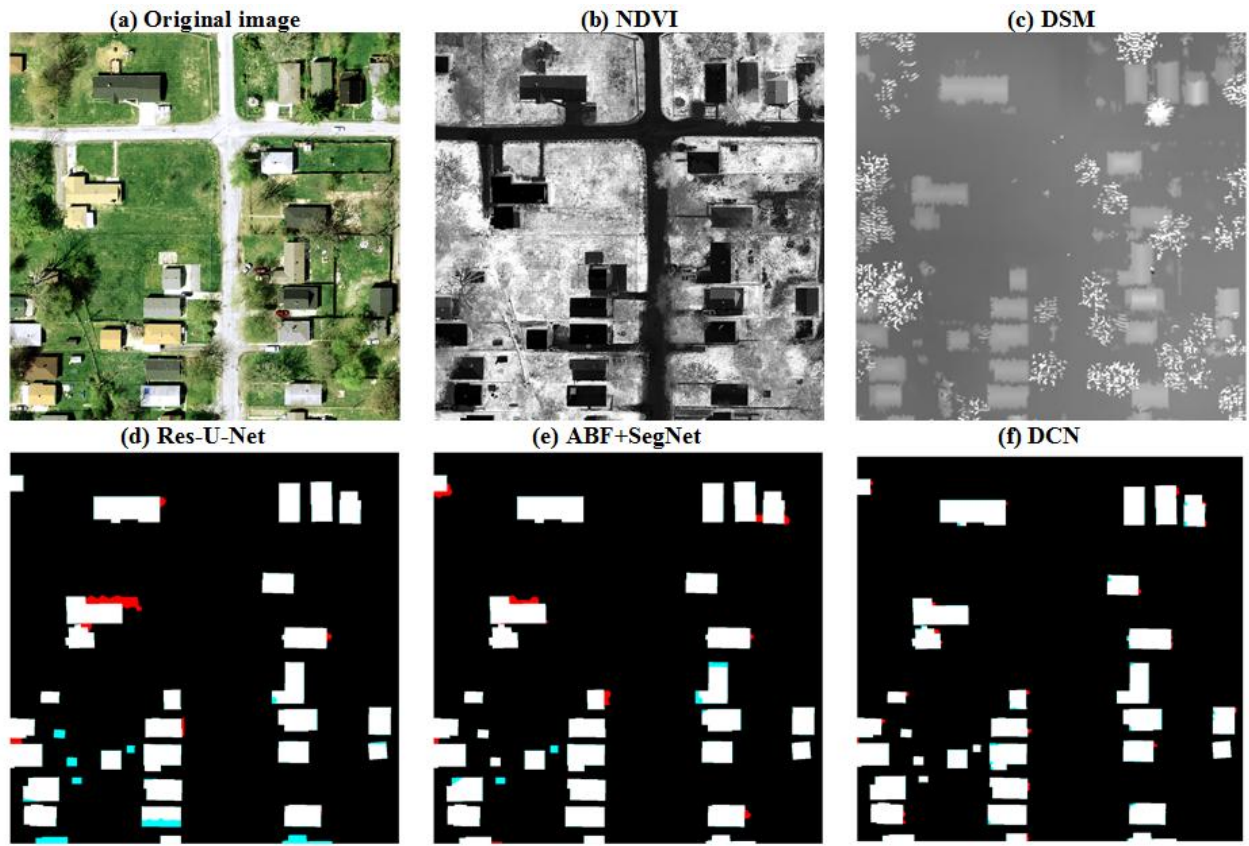

Figure 5. Comparison of BFE results using Res-U-Net, ABF+SegNet and DCN with ground truth in Case-3